\documentclass[runningheads]{llncs}
\usepackage{float}
\usepackage[T1]{fontenc}
\usepackage[hidelinks]{hyperref}
\usepackage{amsmath} %%%%
\usepackage{amssymb} %%%%
\usepackage{amsfonts} %%%%
\usepackage{mathtools} %%%%
\usepackage{bm} %%%%
\usepackage{algorithm}
\usepackage{algpseudocode}
\usepackage{graphicx,verbatim}
\usepackage{multirow}
\usepackage[dvipsnames,table]{xcolor}  % color names + table support
\newcommand{\milmethod}[1]{\textcolor{MidnightBlue}{#1}}   % MIL
\newcommand{\ocmethod}[1]{\textcolor{OrangeRed}{#1}}       % one-class
\newcommand{\dfourmethod}[1]{\textcolor{ForestGreen}{#1}}    % non-equiv. generative
\newcommand{\genmethod}[1]{\textcolor{RoyalPurple}{#1}}  % D4-equivariant
\begin{document}
\title{Group Equivariant Diffusion for Anomaly Detection in Computational Cytology}
\titlerunning{Equivariant Diffusion for Anomaly Detection in Cytology}
% \begin{comment}  %% Removed for anonymized MICCAI submission
% \author{Swarnadip Chatterjee\inst{1}\orcidID{0000-1111-2222-3333} \and
% Ssharvien Kumar Sivakumar\inst{2}\orcidID{1111-2222-3333-4444} \and
% Anirban Mukhopadhyay\inst{2}\orcidID{2222-3333-4444-5555}}
% %
\author{
Swarnadip Chatterjee\inst{1} \and
Ssharvien Kumar Sivakumar\inst{2} \and
Anirban Mukhopadhyay\inst{2}
}

\authorrunning{S. Chatterjee et al.}

\institute{
Uppsala University, Uppsala, Sweden\\
\email{swarnadip.chatterjee@it.uu.se}
\and
Technical University of Darmstadt, Darmstadt, Germany
}

% \author{Anonymized Authors}  %% Added for anonymized MICCAI submission
% \authorrunning{Anonymized Author et al.}
% \institute{Anonymized Affiliations \\
%     \email{email@anonymized.com}}
  
\maketitle              % typeset the header of the contribution
\begin{abstract}
% The abstract should briefly summarize the contents of the paper in 150-250 words.  If you are to include a link to your Repository, please make sure it is anonymized for the double-blind review phase.
Computational cytology on whole-slide images is challenging because malignant cells are rare, heterogeneous, and annotated slides are scarce. Anomaly detection frameworks can be trained on normal slide-negative
patches and then applied at test time to flag abnormal
patches in held-out slides. Most unsupervised anomaly detection approaches including generative ones (GAN-based and diffusion-based), are tuned to organ-level imaging and require large curated datasets. In cytology the signal is cell-centric: rotating or flipping a single-cell patch does not change its diagnostic class, yet standard diffusion models treat transformed views as distinct inputs, leading to transformation-dependent reconstructions and unstable anomaly scores. We propose a $D_4$-equivariant diffusion framework that enforces rotation and reflection symmetry both architecturally, via a $D_4$-equivariant U-Net, and at inference, via equivariant noise coupling and (optionally) frame averaging. This alignment with biological invariance yields transformation-consistent pseudo-healthy reconstructions and more stable anomaly ranking under symmetry. On two publicly available cytology datasets of bone marrow and peripheral blood smears, our $D_4$-equivariant diffusion models achieve higher AUC and retrieve more abnormal cells in the top $K$ predictions than non-equivariant generative baselines, a deep one-class, and a multiple instance learning based method, while substantially reducing score variance across rotations and flips. Code is available at \url{https://swchmida.github.io/D4diffCyto/}.

% Cytology is cell-centric
% rather than organ-centric.At the single-cell level, in-plane rotations and flips do not change the underlying cell type, so labels should be invariant to these transformations. Yet existing generative anomaly detectors do not utilize these symmetries; therefore, we propose a dihedral ($D_4$)-equivariant diffusion framework for data-efficient anomaly detection in computational cytology. Following a partial-diffusion reconstruction strategy, we enforce $D_4$ symmetry both (i) architecturally, via a $D_4$-equivariant U-Net denoiser, and (ii) at inference, via frame-averaging and equivariant noise coupling to produce rotation and flip-consistent reconstructions and scores. We evaluate on two datasets of single-cell patches from bone marrow and blood smears, where we define binary normal-abnormal splits from the provided labels. We primarily compare against generative anomaly detectors (diffusion and GAN based), a one-class baseline, and a MIL framework for patch-based scoring. Results show that $D_4$ equivariant diffusion stabilizes anomaly scores and yields higher AUC and top-K retrievals on the test set when anomaly scores are used for cell-level classification.
\end{abstract}

% Authors must provide keywords and are not allowed to remove this Keyword section.
%
%
%

\section{Introduction}

Cytology \cite{al2011basics} is a core component of pathology in which decisions hinge on subtle single-cell morphology, while truly malignant or severely dysplastic cells can be orders of magnitude rarer than normal or reactive cells, even on malignant slides, leading to extreme class imbalance. Digital cytology and whole-slide scanners now enable high-resolution digitization of entire slides \cite{jiang2023deep,landau2019artificial}, making exhaustive manual review unrealistic when most slide-negative (and many slide-positive) cases are dominated by non-malignant morphology \cite{bengtsson2014screening,cheng2021robust,coustan2011new,kruse2020minimal}. Self- and weakly supervised multiple instance learning (MIL) based slide classifiers \cite{ilse2018attention,li2021dual,liu2024pamil,liu2023multiple,lu2021data,schirris2022deepsmile,torpey2024deepset} can predict slide labels but often fail to reliably surface the few key malignant instances, motivating unsupervised anomaly-detection methods \cite{fernando2021deep,ruff2021unifying} trained on guaranteed-normal slide-negative patches to model normal-cell distributions and flag outliers as candidate abnormal cells \cite{chatterjee2024detection}.

\begin{figure}[t]
    \centering
    \includegraphics[width=\textwidth]{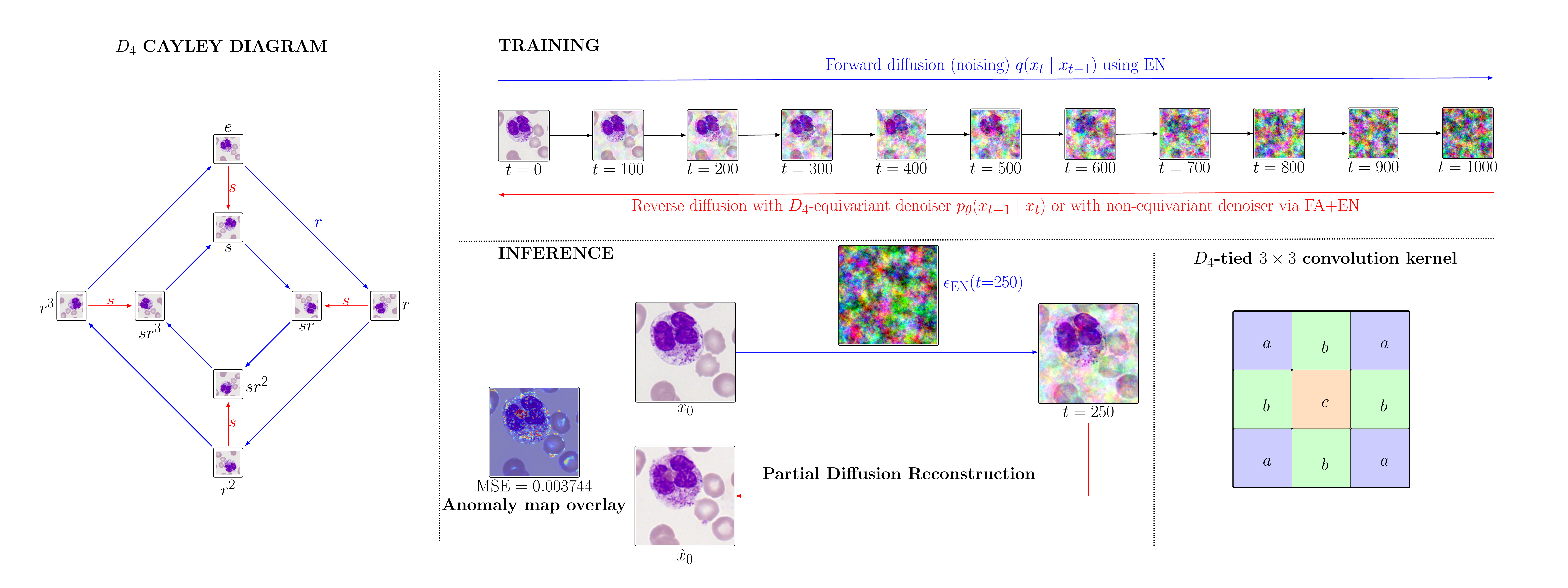}
    \caption{Overview of the proposed $D_4$-equivariant diffusion framework.
    Left: Cayley diagram of the dihedral group $D_4$ acting on a single-cell
    patch. Top-right: training with a $D_4$-equivariant U-Net with
    equivariant noise (EN) coupling. Bottom-right: Inference using partial diffusion reconstruction at test time and an example $D_4$-tied convolution kernel.}
    \label{fig:d4_pipeline}
\end{figure}

A broad range of medical anomaly-detection methods \cite{fernando2021deep} has been explored, from deep one-class models such as Deep SVDD \cite{ruff2018deep} to generative approaches that reconstruct toward a normal manifold and use reconstruction-based scores (e.g., GAN-based \cite{schlegl2019f} and diffusion-based \cite{melba:2025:030:bercea,wolleb2022diffusion}). Diffusion anomaly detectors built on partial diffusion (AnoDDPM-style) \cite{linmans2024diffusion,wyatt2022anoddpm} and their refinements \cite{beizaee2025mad,bercea2023mask,bercea2024diffusion,fuchs2024harp,naval2024ensembled,wolleb2022diffusion} achieve strong lesion localization in organ-centric imaging (brain MRI, retinal OCT, X-ray), where anomalies form contiguous lesions in a fixed anatomy. In cytology, by contrast, abnormalities are local nuclear or nuclear-to-cytoplasmic changes at the single-cell level, so orientation does not change the diagnostic class, yet standard generative detectors treat rotated/flipped views as distinct inputs, increasing sample complexity and inducing orientation-dependent score variation. Under extreme class imbalance and budgeted review, cytologists inspect only a small top-\(K\) set; if rotations or flips change the rank ordering of anomaly scores, the system fails operationally even when global AUC is high. We therefore aim for \emph{consistent ranking under \(D_4\) transforms}, encoding group symmetry to remove nuisance variance from the generative model and stabilize anomaly scores under extreme imbalance.

We propose a dihedral-equivariant diffusion framework that enforces $D_4$
symmetry throughout partial-diffusion reconstruction \cite{lu2025diffusion,zhou2024denoising} (see Fig.~\ref{fig:d4_pipeline}). Concretely, we equip the U\hbox{-}Net denoiser with a $D_4$-equivariant architecture \cite{chidester2019enhanced,lafarge2021roto}, sharing parameters across rotated and flipped views to reduce hypothesis space and redundant variability, and we enforce symmetry at inference via \emph{equivariant noise coupling} (EN) and \emph{frame averaging} (FA) \cite{lu2025diffusion,zhou2024denoising}, which together promote pathwise equivariant trajectories and symmetrized reconstructions. These complementary routes: architectural equivariance and inference-induced equivariance (FA+EN), directly target $D_4$-consistent anomaly scores, and on two publicly available cytology datasets of single-cell patches from bone marrow and peripheral blood smears our $D_4$-equivariant diffusion yields more stable scores across orientations and higher test AUC than non-equivariant generative detectors, a deep one-class baseline, and a MIL-based framework.

% We propose a dihedral-equivariant diffusion framework that enforces $D_4$
% symmetry throughout partial-diffusion reconstruction \cite{lu2025diffusion,zhou2024denoising}(see Fig.~\ref{fig:d4_pipeline}). Concretely, we (i) deploy the standard U-Net denoiser equipped with a $D_4$-equivariant architecture \cite{chidester2019enhanced,lafarge2021roto}, so denoising shares parameters across the full orbit of rotated/flipped views, reducing the effective hypothesis space and the burden of modeling redundant transformation variability. We further (ii) enforce symmetry at inference via \emph{equivariant noise coupling} (EN) \cite{lu2025diffusion,zhou2024denoising}, which couples forward and (optional) reverse stochasticity across group actions to obtain \emph{pathwise} equivariant trajectories, and via \emph{frame averaging} (FA) \cite{lu2025diffusion,zhou2024denoising}, which symmetrizes reconstructions (and hence scores) even when equivariance is only approximate in practice. These two routes: architectural equivariance and inference-induced equivariance (FA$+$EN), are complementary and directly target rotation and flip-consistent reconstructions and anomaly scores. On two publicly available cytology datasets of single-cell patches from bone marrow and peripheral blood smears, this $D_4$-equivariant diffusion approach yields more stable anomaly scores across orientations, and achieves higher test AUC than non-equivariant generative detectors, a deep one-class baseline, and a MIL-based scoring framework.

\section{Method}
\label{sec:method}
In this section, we use diffusion models for anomaly detection on cytology patches and show how we inject $D_4$-equivariance. A denoising diffusion model is trained on slide-negative single-cell patches to learn a normal appearance prior. At test time, we run partial diffusion: each query patch is noised to an intermediate step and then denoised back; the pixelwise residual is the anomaly map and its mean-squared energy is the anomaly score. Standard data augmentation is not enough here: the stochastic reconstruction process can still make scores change under rotations/flips, so our goal is \emph{ranking stability under symmetry}, where patches differing only by $D_4$ transforms receive comparable scores.

We pursue two routes to $D_4$-consistent scoring, both covered by Algorithm~\ref{alg:d4diff_unified}. First, we use a standard (non-equivariant) U\hbox{-}Net trained with the usual diffusion loss and impose $D_4$ structure only at inference via \emph{equivariant noise} (EN) and \emph{frame averaging} (FA) \cite{lu2025diffusion} over rotated and flipped views.
Second, we use a $D_4$-equivariant U\hbox{-}Net, built from group-equivariant convolutions and, in one variant, a $D_4$-equivariant attention block with shared group-transformed relative positional biases; this backbone is trained with the same objective, and at test time Algorithm~\ref{alg:d4diff_unified} is run with EN-only, letting architectural equivariance carry the $D_4$ structure through the sampling trajectory.

\begin{algorithm}[t]
\caption{$D_4$-aware partial-diffusion anomaly detection}
\label{alg:d4diff_unified}
\begin{algorithmic}[1]
\State \textbf{Inputs:} normal training set $\{x_0^{(i)}\}_{i=1}^N$, test patch $x_0$, steps $T$, partial step $t^\star$, group $G=D_4$,
\Statex \hspace{\algorithmicindent} backbone $f_\theta \in \{\mathrm{UNet},\mathrm{UNet}_{D_4}\}$, flags EN, FA
\vspace{0.3em}
\State \textbf{Training}
\For{minibatch $\mathcal{B}$}
  \State draw $t \sim \mathcal{U}\{1,\dots,T\}$, $\epsilon \sim \mathcal{N}(0,\mathbf{I})$; 
         set $x_t = \sqrt{\bar{\alpha}_t}x_0 + \sqrt{1-\bar{\alpha}_t}\epsilon$ for $x_0\in\mathcal{B}$
  \State $\hat{\epsilon} = f_\theta(x_t,t)$; 
         $\mathcal{L} = \|\epsilon - \hat{\epsilon}\|_2^2$; 
         update $\theta \leftarrow \theta - \eta\nabla_\theta \mathcal{L}$
\EndFor
\vspace{0.3em}
\State \textbf{Inference (score for $x_0$)}
\If{FA}
  \State $\mathcal{O}(x_0)=\{g\cdot x_0 : g\in G\}$; draw $\epsilon,\{\eta_t\}_{t=1}^{t^\star}$
  \For{$g\in G$}
    \State \textbf{if} EN \textbf{then} $\epsilon_g=g\cdot\epsilon,\,\eta_{t,g}=g\cdot\eta_t$; \textbf{else} draw $\epsilon_g,\eta_{t,g}$
    \State $x_{t^\star}^g=\mathrm{DiffForward}(g\cdot x_0,\epsilon_g,t^\star)$; 
           $\hat{x}_0^{\,g}=\mathrm{DiffReverse}(x_{t^\star}^g,\eta_{1:t^\star,g})$
    \State $A^g=|g\cdot x_0-\hat{x}_0^{\,g}|$; \quad 
           $S^g=\tfrac{1}{CHW}\|g\cdot x_0-\hat{x}_0^{\,g}\|_2^2$
  \EndFor
  \State $A^{\mathrm{FA}}(x_0)=\tfrac{1}{|G|}\sum_{g\in G} g^{-1}\!\cdot A^g$; \quad
         $S^{\mathrm{FA}}(x_0)=\tfrac{1}{|G|}\sum_{g\in G} S^g$; \quad
         \textbf{return} $S^{\mathrm{FA}}(x_0)$
\Else
  \State draw $\epsilon,\{\eta_t\}_{t=1}^{t^\star}$; 
         $x_{t^\star}=\mathrm{DiffForward}(x_0,\epsilon,t^\star)$;
         $\hat{x}_0=\mathrm{DiffReverse}(x_{t^\star},\eta_{1:t^\star})$
  \State $A(x_0)=|x_0-\hat{x}_0|$; \quad
         $S(x_0)=\tfrac{1}{CHW}\|x_0-\hat{x}_0\|_2^2$; \quad
         \textbf{return} $S(x_0)$
\EndIf
\end{algorithmic}
\end{algorithm}

\subsection{Diffusion anomaly detection via partial reconstruction}
\label{sec:method:diffusion_ad}

Let $x_0 \in \mathbb{R}^{C \times H \times W}$ denote a single-cell patch, with training samples drawn from the normal distribution $p_n$ and test samples drawn from held-out slides. We use a standard denoising diffusion model: the forward process adds Gaussian noise in $T$ steps via $q(x_t \mid x_0)=\mathcal{N}(\sqrt{\bar{\alpha}_t}\,x_0,(1-\bar{\alpha}_t)\mathbf{I})$, and the denoiser $f_\theta(x_t,t)$ is trained with the usual DDPM noise-prediction loss $\mathcal{L}(\theta)=\mathbb{E}_{x_0,t,\epsilon}\big[\lVert \epsilon - f_\theta(x_t,t)\rVert_2^2\big]$ on slide-negative patches. 

At test time we run \emph{partial diffusion}: for a chosen step $t^\star$, we diffuse $x_0$ to $x_{t^\star}$ using the forward process and then apply $t^\star$ reverse steps (Algorithm~\ref{alg:d4diff_unified}) to obtain a pseudo-normal reconstruction $\hat{x}_0=\mathcal{R}(x_0;\epsilon,\eta_{1:t^\star})$. We then define an anomaly map and scalar anomaly score as $A(x_0)=|x_0-\hat{x}_0|$ and $S(x_0)=\frac{1}{CHW}\,\|x_0-\hat{x}_0\|_2^2$. All diffusion-based variants in this work use $S(x_0)$ as the final anomaly score; $A(x_0)$ is used only for qualitative visualization.

\subsection{$D_4$-equivariant sampling and architectures}
\label{sec:method:d4}

We model in-plane rotations and flips of single-cell patches using the dihedral group $G=D_4$, acting on images by pixel re-indexing $(g\cdot x)(u)=x(g^{-1}u)$. Our goal is for the scalar score $S(\cdot)$ to be approximately invariant under $G$, so that patches differing only by $D_4$ transforms receive comparable anomaly scores.

To couple stochasticity across the orbit, we use \emph{equivariant noise} (EN): a single base noise sample $(\epsilon,\{\eta_t\}_{t=1}^{t^\star})$ is transformed as $\epsilon_g = g\cdot\epsilon$ and $\eta_{t,g}=g\cdot\eta_t$ for each $g\in G$, so the forward and reverse trajectories for $g\cdot x_0$ share the same underlying randomness as for $x_0$, up to the group action. When desired, we further apply \emph{frame averaging} (FA), averaging residuals and scores over the $D_4$ orbit to stabilize $A(x_0)$ and $S(x_0)$.

Architectural and sampling choices are combined in a unified way: Algorithm~\ref{alg:d4diff_unified} toggles (i) the backbone (standard vs.\ $D_4$-equivariant U\hbox{-}Net) and (ii) the EN/FA flags. For a generic U\hbox{-}Net we use EN+FA at inference to enforce $D_4$ consistency, while for the $D_4$-equivariant U\hbox{-}Net (with group-tied convolutions and, in our second variant, $D_4$-equivariant attention) we typically use EN-only, relying on architectural equivariance to propagate group structure through the sampling trajectory.

\subsection{Theoretical guarantees}
\label{sec:method:theory}

In our setting, $D_4$ invariance refers to class invariance of centered single-cell patches: rotating or flipping a cell patch should not change its diagnostic class. 

\subsubsection{D\texorpdfstring{\boldmath$_4$}{4}-averaging cannot increase DDPM risk}
\label{sec:method:train_symm}

We consider the idealized setting where $X_0\sim p_n$ is $D_4$-invariant and the injected noise $E\sim \mathcal{N}(0,\mathbf{I})$ is isotropic. Let $X_t$ be the noised samples, $E$ be the sampled equivariant noise and assume the training pairs are $D_4$-invariant in distribution:
\( (X_t,E)\overset{d}{=}(g\cdot X_t,\; g\cdot E)\qquad \forall g\in G.
\label{eq:pair_invariance} \)
Define the $D_4$-averaged denoiser
\( f_\theta^{\mathrm{avg}}(x_t,t) = \frac{1}{|G|}\sum_{g\in G} g^{-1}\!\cdot f_\theta(g\cdot x_t, t). \label{eq:sym_denoiser} \)
For the squared-error loss $\ell(\hat\epsilon,\epsilon)=\|\hat\epsilon-\epsilon\|_2^2$, Jensen’s inequality together with $D_4$ pair invariance yields
\( \mathbb{E}\big[\ell(f_\theta^{\mathrm{avg}}(X_t,t),E)\big]
\le
\mathbb{E}\big[\ell(f_\theta(X_t,t),E)\big].
\label{eq:risk_sym_leq} \)
Thus, symmetrizing the denoiser with respect to $D_4$ cannot increase the DDPM training risk and can, in principle, improve it by averaging out orientation-specific fluctuations.

\noindent\textbf{Invariant detection under class-conditional symmetry:} Let $Y\in\{0,1\}$ denote normal/abnormal. Assume the class-conditionals are $G$-invariant:
\( p(x\mid y)=p(g\cdot x\mid y) \forall g\in G,\; y\in\{0,1\}.
\label{eq:class_cond_invariance} \)
For any score $s(x)$, define its group-averaged version $\bar{s}(x)=\frac{1}{|G|}\sum_{g\in G} s(g\cdot x)$. For a false-positive budget $\alpha$, the optimal miss-rate among threshold rules based on $s$ is
\( \beta^\star(\alpha; s)
=\inf_{\tau}\; \Pr\!\big(s(X)\le \tau \mid Y=1\big)
\;\;\text{s.t.}\;\;
\Pr\!\big(s(X)>\tau \mid Y=0\big)\le \alpha .
\label{eq:beta_star} \)
Under $G$-invariant class-conditionals, restricting to $G$-invariant scores (e.g., scores obtained by averaging over $D_4$ orbits) does not worsen this optimal trade-off. In other words, symmetry-respecting averaging is theoretically justified when the underlying data distribution exhibits the corresponding rotational and flip symmetries, aligning our $D_4$-equivariant diffusion design with the structure of single-cell cytology images.

\section{Experiments and Results}
\label{sec:experiments}

In this section, we describe the datasets and experimental setup, then report quantitative and qualitative results for our $D_4$-equivariant diffusion models and competing baselines, followed by targeted ablation studies.

\subsection{Datasets and splits}

We evaluate on two cytology single–cell datasets.
The MLL bone marrow dataset \cite{matek2019_aml_lmu_dataset} contains $250{\times}250$ RGB patches from bone marrow smears. We treat typical lymphocytes (LYT) as the normal class, using $9{,}185$ LYT patches for training, $7{,}873$ LYT patches for
testing, and merge all seven minority diagnostic subsets
(e.g.\ blasts and other rare morphologies) into a single
abnormal class with $396$ test patches.

The AML LMU Cytomorphology dataset \cite{matek2021_bm_cytology_dataset} contains $400{\times}400$ RGBA patches from peripheral blood smears. We group mature leukocytes as normal and immature leukocytes or \textit{other} types as abnormal.
From the mature group, we use $10{,}383$ patches for training and $4{,}450$ for testing; from the immature group we use all
$3{,}532$ patches for testing.

\begin{table}
\centering
\caption{Patch-level anomaly detection on MLL bone marrow and AML Cytomorphology LMU datasets (AUC and number of true positives among top-\(K\) ranked patches).}
\label{tab:auc_both}
\begin{tabular}{|l|c|ccc|c|ccc|}
\hline
\multirow{3}{*}{\textbf{Method}} &
\multicolumn{4}{c|}{\textbf{MLL}} &
\multicolumn{4}{c|}{\textbf{AML LMU}} \\
\cline{2-9}
& \textbf{AUC} & \multicolumn{3}{c|}{\textbf{True Positives}} &
  \textbf{AUC} & \multicolumn{3}{c|}{\textbf{True Positives}} \\
\cline{3-5}\cline{7-9}
& & \textbf{TP\(_{400}\)} & \textbf{TP\(_{200}\)} & \textbf{TP\(_{100}\)} &
  & \textbf{TP\(_{400}\)} & \textbf{TP\(_{200}\)} & \textbf{TP\(_{100}\)} \\
\hline
\multicolumn{9}{|l|}{\texttt{Baselines}} \\
\hline
\milmethod{ItS2CLR}              & 0.521107 &  27 &  20 & 19 & 0.360515 &  92 &  38 & 13 \\
\ocmethod{Deep SVDD}            & 0.613331 &  49 &  35 & 21 & 0.514742 & 227 & 122 & 56 \\
\genmethod{f-AnoGAN}            & 0.594601 &  37 &  24 &  9 & 0.538857 & 273 & 131 & 57 \\
\genmethod{THOR (\texttt{Gaussian})}      & 0.540851 &  24 &  14 &  7 & 0.582212 & 247 & 137 & 68 \\
\genmethod{THOR (\texttt{Simplex})}       & 0.404264 &   8 &   3 &  1 & 0.544415 & 207 & 104 & 58 \\
\genmethod{BerDiff}             & 0.550799 &  44 &  32 & 26 & 0.708558 & 313 & 157 & 64 \\
\genmethod{AnoDDPM}             & 0.632846 &  57 &  30 & 18 & 0.694320 & 311 & \textbf{162} & 83 \\
\hline
\multicolumn{9}{|l|}{\texttt{$D_4$-equivariant diffusion variants}} \\
\hline
\dfourmethod{FA only (\texttt{inference})}  & 0.612543 &  46 &  32 & 19 & 0.680111 & 277 & 144 & 72 \\
\dfourmethod{FA+EN (\texttt{inference})}    & \textbf{0.684121} & \textbf{78} & \textbf{53} & \textbf{34}
                     & 0.678373 & 285 & 149 & 75 \\
\dfourmethod{$D_4$ \texttt{conv.}}          & 0.675404 &  74 & \textbf{53} & 27
                     & 0.699635 & 315 & 159 & \textbf{86} \\
\dfourmethod{$D_4$ \texttt{conv.+attn.}}    & 0.679251 &  67 &  44 & 33
                     & \textbf{0.720098} & \textbf{325} & \textbf{162} & 81 \\
\hline
\end{tabular}
\end{table}

\subsection{Setup and baselines}

We benchmark \emph{patch-level} anomaly detection.
Each method assigns an anomaly score to every test patch and we report
AUC (Area Under the Receiver Operating Characteristic Curve) as the primary metric. Since cytology screening is dominated by extreme class imbalance and review is budgeted, we treat each test set for both datasets as a single abnormal \textit{slide} containing a mix of normal and abnormal single-cell patches. Thus, we also measure the number of true positives in the top-$K$ most anomalous predictions (TP$_{400}$, TP$_{200}$, TP$_{100}$), reflecting realistic pre-screening where only a limited candidate set is inspected. For our AnoDDPM and $D_4$-equivariant AnoDDPM variants, the anomaly score is the mean-squared error (MSE) between the input patch and its reconstruction.

As baselines we include:
(i) a MIL-based instance scorer (ItS2CLR) \cite{liu2023multiple};
(ii) a deep one-class method (Deep SVDD)\cite{ruff2018deep};
and (iii) representative generative detectors:
f-AnoGAN\cite{schlegl2019f}, THOR (with gaussian and simplex noise)\cite{bercea2024diffusion}, Masked Bernoulli Diffusion
(BerDiff)\cite{wolleb2024binary}, and a vanilla AnoDDPM detector\cite{wyatt2022anoddpm} (non-equivariant U-Net,
simplex noise, partial diffusion).
All methods are trained on the same normal training splits and evaluated on exactly the same test sets.

Our $D_4$-equivariant diffusion models fall into two families: those that impose group structure only at inference via Frame Averaging (FA) and Equivariant Noise (EN), and those that build $D_4$ structure into the U\hbox{-}Net backbone; the concrete instantiations are detailed in Section~\ref{subsec:ablation}.

\subsection{Quantitative results}

Table~\ref{tab:auc_both} summarizes AUC and TP\(_K\) on both datasets.
On MLL, the non-equivariant AnoDDPM baseline reaches
AUC \(=0.632\) with TP\(_{400}=57\).
Our best $D_4$ variant, FA+EN with a generic U-Net, improves AUC to
\(0.684\) and TP\(_{400}\) to \(78\), while the $D_4$-equivariant
convolutional U-Net (no attention) yields AUC \(=0.679\) and TP\(_{400}=74\).
This shows that enforcing $D_4$ structure, either architecturally or via
inference-time averaging, leads to both better global separation and
denser concentration of truly abnormal cells at the top of the ranked list.

\begin{figure}
\centering
\includegraphics[width=\textwidth]{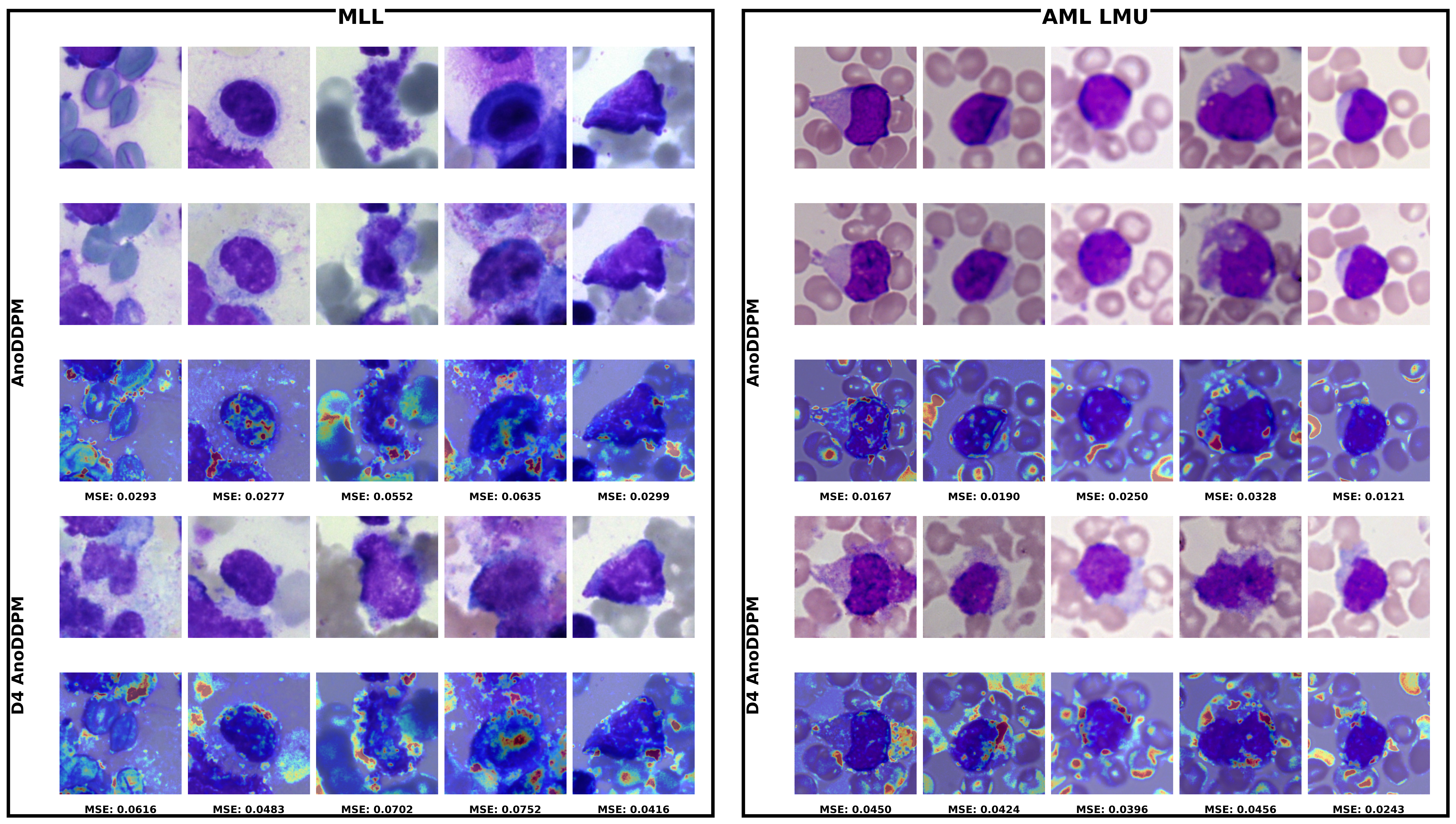}
\caption{Qualitative comparison of AnoDDPM vs.\ $D_4$-equivariant AnoDDPM on
MLL bone marrow (left panel) and AML LMU (right panel).
For each method we show inputs (\textit{row 1}), reconstructions (\textit{rows 2 and 4}), and qualitative reconstruction-deviation overlays with patch-level anomaly-score MSE values (\textit{rows 3 and 5}).}
\label{fig:qualitative}
\end{figure}

On AML LMU, BerDiff and vanilla AnoDDPM are strong generative baselines
(AUC \(=0.708\) and \(0.694\), respectively).
Here the best performance is obtained with the
$D_4$-equivariant U-Net with attention:
AUC \(=0.720\), TP\(_{400}=325\), and TP\(_{200}=162\).
This indicates that $D_4$-equivariant diffusion remains beneficial
in a different cytology modality (blood smears) as well. Figure~\ref{fig:qualitative} shows some qualitative examples on both datasets.

\subsection{Ablation studies}
\label{subsec:ablation}
\noindent{\textbf{Inference-induced vs.\ architectural $D_4$ variants.}}
We ablate four $D_4$ variants (Table~\ref{tab:auc_both}). On MLL, FA-only with a standard U\hbox{-}Net slightly degrades AUC, while FA+EN recovers and improves over vanilla AnoDDPM. The $D_4$ convolutional U\hbox{-}Net on the other hand, slightly outperforms the attention-augmented variant. On AML LMU, FA-only and FA+EN each improve top-\(K\) retrieval over the baseline, whereas the architectural $D_4$ models further strengthen both AUC and TP counts, with the attention-based variant giving the strongest overall diffusion performance.

\noindent{\textbf{Anomaly-score invariance under $D_4$ transforms.}}
On MLL, we sample 100 normal and 100 abnormal patches and generate
their rotations by $90^\circ$, $180^\circ$, $270^\circ$ and
horizontal flips, then measure variation of anomaly scores as
(mean SD, median SD, mean range) per image, where SD is standard deviation.
Vanilla AnoDDPM yields $(0.00375,\;0.00343,\;0.01036)$, whereas the
$D_4$ FA+EN variant reduces this to $\mathbf{(0.00206,\;0.00161,\;0.00515)}$ and the
$D_4$ U\hbox{-}Net without attention to $(0.00224,\;0.00177,\;0.00560)$,
indicating substantially more rotation- and flip-consistent anomaly scores for
the dihedral-aware diffusion models.

\noindent{\textbf{Uncertainty and statistical significance.}}
For the MLL dataset, we used stratified bootstrap resampling of test patches and compared each $D_4$-equivariant model against AnoDDPM via paired two-sided tests on AUC and TP\(_{400}\). AnoDDPM obtained AUC 95\% CI $(0.603,\;0.661)$ and TP\(_{400}\) CI $(43,\;71)$, while the $D_4$ FA+EN variant tightened these to $(0.657,\;0.710)$ and $(61,\;93)$, and the $D_4$ convolutional U\hbox{-}Net (no attention) to $(0.647,\;0.703)$ and $(58,\;89)$. Paired bootstrap tests gave $p<10^{-3}$ for AUC and $\mathbf{p=0.006}\,\, $\textbf{for TP\(_{400}\) for both $D_4$ models versus AnoDDPM}, making the proposed method clinically significant and indicating statistically reliable gains in ranking and top-\(K\) retrieval. On the AML LMU dataset, BerDiff showed AUC CI $(0.697,\;0.720)$ and TP\(_{400}\) CI $(295,\;328)$, whereas the $D_4$-equivariant U\hbox{-}Net with attention achieved $(0.709,\;0.731)$ and $(307,\;340)$. Paired bootstrap tests yielded $p\approx 0.125$ for AUC and $p\approx 0.275$ for TP\(_{400}\), so the $D_4$ model’s advantage is numerically consistent but statistically inconclusive on AML LMU; with the clearly significant MLL gains. Together, these results suggest that enforcing $D_4$ structure can improve anomaly ranking, with strong evidence on MLL and encouraging trends on AML LMU.
\section{Conclusion}

We addressed a structural mismatch between generative anomaly detection and computational cytology. In cell-centric imaging, diagnostic class is invariant to in-plane rotations and reflections, yet standard diffusion models yield transformation-dependent reconstructions and unstable anomaly scores. We introduced a $D_4$-equivariant diffusion framework that enforces dihedral symmetry in both architecture and sampling, yielding rotation/flip-consistent reconstructions and more stable anomaly ranking under extreme class imbalance. Across two cytology datasets, enforcing $D_4$ structure improved global separation and clinically relevant top-$K$ retrieval: on MLL, the $D_4$ FA+EN variant raised $\mathrm{TP}_{400}$ from 57 (AnoDDPM) to 78 with statistically significant gains in both AUC and $\mathrm{TP}_{400}$ $(p = 0.006)$, while on AML LMU, the $D_4$-equivariant U\hbox{-}Net with attention achieved the best AUC (0.720) and $\mathrm{TP}_{400}$ (325). Overall, aligning generative anomaly detection with cytology’s dihedral symmetries enhances ranking stability and improves practical abnormal-cell retrieval in realistic screening workflows.
\\
\\
\\
{\small
\noindent\textbf{Disclosure of Interests.}
The authors have no competing interests to declare that are relevant to the content of this article.
}
\\
\\
{\small
\noindent\textbf{Acknowledgements.}
This work was supported by the Liljewalch Foundation, Sweden, through a travel scholarship awarded to the first author for a research visit to Technical University of Darmstadt.
}

\bibliographystyle{splncs04}
\bibliography{Paper-2803}

@article{al2011basics,
  title={Basics of cytology},
  author={Al-Abbadi, Mousa A},
  journal={Avicenna journal of medicine},
  volume={1},
  number={01},
  pages={18--28},
  year={2011},
  publisher={Thieme Medical and Scientific Publishers Private Ltd.}
}

@article{jiang2023deep,
  title={Deep learning for computational cytology: A survey},
  author={Jiang, Hao and Zhou, Yanning and Lin, Yi and Chan, Ronald CK and Liu, Jiang and Chen, Hao},
  journal={Medical Image Analysis},
  volume={84},
  pages={102691},
  year={2023},
  publisher={Elsevier}
}

@article{landau2019artificial,
  title={Artificial intelligence in cytopathology: a review of the literature and overview of commercial landscape},
  author={Landau, Michael S and Pantanowitz, Liron},
  journal={Journal of the American Society of Cytopathology},
  volume={8},
  number={4},
  pages={230--241},
  year={2019},
  publisher={Elsevier}
}

@article{kruse2020minimal,
  title={Minimal residual disease detection in acute lymphoblastic leukemia},
  author={Kruse, Aaron and Abdel-Azim, Nour and Kim, Hye Na and Ruan, Yongsheng and Phan, Valerie and Ogana, Heather and Wang, William and Lee, Rachel and Gang, Eun Ji and Khazal, Sajad and others},
  journal={International journal of molecular sciences},
  volume={21},
  number={3},
  pages={1054},
  year={2020},
  publisher={MDPI}
}

@article{coustan2011new,
  title={New markers for minimal residual disease detection in acute lymphoblastic leukemia},
  author={Coustan-Smith, Elaine and Song, Guangchun and Clark, Christopher and Key, Laura and Liu, Peixin and Mehrpooya, Mohammad and Stow, Patricia and Su, Xiaoping and Shurtleff, Sheila and Pui, Ching-Hon and others},
  journal={Blood, The Journal of the American Society of Hematology},
  volume={117},
  number={23},
  pages={6267--6276},
  year={2011},
  publisher={American Society of Hematology Washington, DC}
}

@article{bengtsson2014screening,
  title={Screening for cervical cancer using automated analysis of PAP-smears},
  author={Bengtsson, Ewert and Malm, Patrik},
  journal={Computational and mathematical methods in medicine},
  volume={2014},
  number={1},
  pages={842037},
  year={2014},
  publisher={Wiley Online Library}
}

@article{cheng2021robust,
  title={Robust whole slide image analysis for cervical cancer screening using deep learning},
  author={Cheng, Shenghua and Liu, Sibo and Yu, Jingya and Rao, Gong and Xiao, Yuwei and Han, Wei and Zhu, Wenjie and Lv, Xiaohua and Li, Ning and Cai, Jing and others},
  journal={Nature communications},
  volume={12},
  number={1},
  pages={5639},
  year={2021},
  publisher={Nature Publishing Group UK London}
}

@article{schirris2022deepsmile,
  title={DeepSMILE: Contrastive self-supervised pre-training benefits MSI and HRD classification directly from H\&E whole-slide images in colorectal and breast cancer},
  author={Schirris, Yoni and Gavves, Efstratios and Nederlof, Iris and Horlings, Hugo Mark and Teuwen, Jonas},
  journal={Medical image analysis},
  volume={79},
  pages={102464},
  year={2022},
  publisher={Elsevier}
}

@article{torpey2024deepset,
  title={DeepSet SimCLR: Self-supervised deep sets for improved pathology representation learning},
  author={Torpey, David and Klein, Richard},
  journal={Pattern Recognition Letters},
  volume={186},
  pages={64--70},
  year={2024},
  publisher={Elsevier}
}

@inproceedings{liu2023multiple,
  title={Multiple instance learning via iterative self-paced supervised contrastive learning},
  author={Liu, Kangning and Zhu, Weicheng and Shen, Yiqiu and Liu, Sheng and Razavian, Narges and Geras, Krzysztof J and Fernandez-Granda, Carlos},
  booktitle={IEEE CVPR},
  pages={3355--3365},
  year={2023}
}

@article{ruff2021unifying,
  title={A unifying review of deep and shallow anomaly detection},
  author={Ruff, Lukas and Kauffmann, Jacob R and Vandermeulen, Robert A and Montavon, Gr{\'e}goire and Samek, Wojciech and Kloft, Marius and Dietterich, Thomas G and M{\"u}ller, Klaus-Robert},
  journal={Proceedings of the IEEE},
  volume={109},
  number={5},
  pages={756--795},
  year={2021},
  publisher={IEEE}
}

@article{fernando2021deep,
  title={Deep learning for medical anomaly detection--a survey},
  author={Fernando, Tharindu and Gammulle, Harshala and Denman, Simon and Sridharan, Sridha and Fookes, Clinton},
  journal={ACM Computing Surveys (CSUR)},
  volume={54},
  number={7},
  pages={1--37},
  year={2021},
  publisher={ACM New York, NY, USA}
}

@inproceedings{chatterjee2024detection,
  title={Detection of Extremely Sparse Key Instances in Whole Slide Cytology Images via Self-supervised One-class Representation Learning},
  author={Chatterjee, Swarnadip and G{\"o}ksel, Orcun and Sladoje, Nata{\v{s}}a and Lindblad, Joakim},
  booktitle={ICPR},
  pages={408--421},
  year={2024},
  organization={Springer}
}

@inproceedings{ilse2018attention,
  title={Attention-based deep multiple instance learning},
  author={Ilse, Maximilian and Tomczak, Jakub and Welling, Max},
  booktitle={ICML},
  pages={2127--2136},
  year={2018},
  organization={PMLR}
}

@article{lu2021data,
  title={Data-efficient and weakly supervised computational pathology on whole-slide images},
  author={Lu, Ming Y and Williamson, Drew FK and Chen, Tiffany Y and Chen, Richard J and Barbieri, Matteo and Mahmood, Faisal},
  journal={Nature biomedical engineering},
  volume={5},
  number={6},
  pages={555--570},
  year={2021},
  publisher={Nature Publishing Group UK London}
}

@inproceedings{li2021dual,
  title={Dual-stream multiple instance learning network for whole slide image classification with self-supervised contrastive learning},
  author={Li, Yuexiang and Liu, Yunzhi and Xu, Yufei and Zhang, Liangqiong and Xing, Lei and Huang, Junzhou},
  booktitle={IEEE ICCV},
  pages={14318--14327},
  year={2021}
}

@inproceedings{liu2024pamil,
  title={Pamil: Prototype attention-based multiple instance learning for whole slide image classification},
  author={Liu, Jiashuai and Mao, Anyu and Niu, Yi and Zhang, Xianli and Gong, Tieliang and Li, Chen and Gao, Zeyu},
  booktitle={MICCAI},
  pages={362--372},
  year={2024},
  organization={Springer}
}

@inproceedings{ruff2018deep,
  title={Deep one-class classification},
  author={Ruff, Lukas and Vandermeulen, Robert A and G{\"o}rnitz, Nico and Deecke, Lucas and Siddiqui, Shoaib A and Binder, Alexander and M{\"u}ller, Emmanuel and Kloft, Marius},
  booktitle={ICML},
  pages={4393--4402},
  year={2018},
  organization={PMLR}
}

@article{schlegl2019f,
  title={f-AnoGAN: Fast unsupervised anomaly detection with generative adversarial networks},
  author={Schlegl, Thomas and Seeb{\"o}ck, Philipp and Waldstein, Sebastian M and Langs, Georg and Schmidt-Erfurth, Ursula},
  journal={Medical image analysis},
  volume={54},
  pages={30--44},
  year={2019},
  publisher={Elsevier}
}

@inproceedings{wolleb2022diffusion,
  title={Diffusion models for medical anomaly detection},
  author={Wolleb, Julia and Bieder, Florentin and Sandk{\"u}hler, Robin and Cattin, Philippe C},
  booktitle={MICCAI},
  pages={35--45},
  year={2022},
  organization={Springer}
}

@article{melba:2025:030:bercea,
    title = "Denoising Diffusion Models for Anomaly Localization in Medical Images",
    author = "Bercea, Cosmin I. and Cattin, Philippe C. and Schnabel, Julia A. and Wolleb, Julia",
    journal = "Machine Learning for Biomedical Imaging",
    year = "2025"
}

@inproceedings{wyatt2022anoddpm,
  title={Anoddpm: Anomaly detection with denoising diffusion probabilistic models using simplex noise},
  author={Wyatt, Julian and Leach, Adam and Schmon, Sebastian M and Willcocks, Chris G},
  booktitle={IEEE CVPR},
  pages={650--656},
  year={2022}
}

@article{linmans2024diffusion,
  title={Diffusion models for out-of-distribution detection in digital pathology},
  author={Linmans, Jasper and Raya, Gabriel and van der Laak, Jeroen and Litjens, Geert},
  journal={Medical Image Analysis},
  volume={93},
  pages={103088},
  year={2024},
  publisher={Elsevier}
}

@inproceedings{wolleb2024binary,
  title={Binary noise for binary tasks: Masked bernoulli diffusion for unsupervised anomaly detection},
  author={Wolleb, Julia and Bieder, Florentin and Friedrich, Paul and Zhang, Peter and Durrer, Alicia and Cattin, Philippe C},
  booktitle={MICCAI},
  pages={135--145},
  year={2024},
  organization={Springer}
}

@inproceedings{beizaee2025mad,
  title={Mad-ad: Masked diffusion for unsupervised brain anomaly detection},
  author={Beizaee, Farzad and Lodygensky, Gregory and Desrosiers, Christian and Dolz, Jose},
  booktitle={IPMI},
  pages={139--153},
  year={2025},
  organization={Springer}
}

@inproceedings{bercea2024diffusion,
  title={Diffusion models with implicit guidance for medical anomaly detection},
  author={Bercea, Cosmin I and Wiestler, Benedikt and Rueckert, Daniel and Schnabel, Julia A},
  booktitle={MICCAI},
  pages={211--220},
  year={2024},
  organization={Springer}
}

@article{bercea2023mask,
  title={Mask, stitch, and re-sample: Enhancing robustness and generalizability in anomaly detection through automatic diffusion models},
  author={Bercea, Cosmin I and Neumayr, Michael and Rueckert, Daniel and Schnabel, Julia A},
  journal={arXiv preprint arXiv:2305.19643},
  year={2023}
}

@inproceedings{naval2024ensembled,
  title={Ensembled cold-diffusion restorations for unsupervised anomaly detection},
  author={Naval Marimont, Sergio and Siomos, Vasilis and Baugh, Matthew and Tzelepis, Christos and Kainz, Bernhard and Tarroni, Giacomo},
  booktitle={MICCAI},
  pages={243--253},
  year={2024},
  organization={Springer}
}

@inproceedings{lu2025diffusion,
  title={Diffusion Models under Group Transformations},
  author={Lu, Haoye and Szabados, Spencer and Yu, Yaoliang},
  booktitle={AISTATS},
  year={2025}
}

@inproceedings{
zhou2024denoising,
title={Denoising Diffusion Bridge Models},
author={Linqi Zhou and Aaron Lou and Samar Khanna and Stefano Ermon},
booktitle={ICLR},
year={2024},
url={https://openreview.net/forum?id=FKksTayvGo}
}

@inproceedings{chidester2019enhanced,
  title={Enhanced rotation-equivariant u-net for nuclear segmentation},
  author={Chidester, Benjamin and Ton, That-Vinh and Tran, Minh-Triet and Ma, Jian and Do, Minh N},
  booktitle={IEEE CVPR Workshops},
  year={2019}
}

@article{lafarge2021roto,
  title={Roto-translation equivariant convolutional networks: Application to histopathology image analysis},
  author={Lafarge, Maxime W and Bekkers, Erik J and Pluim, Josien PW and Duits, Remco and Veta, Mitko},
  journal={Medical Image Analysis},
  volume={68},
  pages={101849},
  year={2021},
  publisher={Elsevier}
}

@misc{matek2019_aml_lmu_dataset,
  author       = {Matek, Christian and Schwarz, Sebastian and Marr, Carsten and Spiekermann, Karsten},
  title        = {A Single-cell Morphological Dataset of Leukocytes from AML Patients and Non-malignant Controls},
  year         = {2019},
  howpublished = {The Cancer Imaging Archive (TCIA) [Data set]}
}

@misc{matek2021_bm_cytology_dataset,
  author       = {Matek, Christian and Krappe, Sebastian and M{\"u}nzenmayer, Christian and Haferlach, Torsten and Marr, Carsten},
  title        = {An Expert-Annotated Dataset of Bone Marrow Cytology in Hematologic Malignancies},
  year         = {2021},
  howpublished = {The Cancer Imaging Archive (TCIA) [Data set]}
}

@inproceedings{fuchs2024harp,
  title={Harp: Unsupervised histopathology artifact restoration},
  author={Fuchs, Moritz and Sivakumar, Ssharvien Kumar R and Sch{\"o}ber, Mirko and Woltering, Niklas and Eich, Marie-Lisa and Schweizer, Leonille and Mukhopadhyay, Anirban},
  booktitle={Medical Imaging with Deep Learning},
  pages={465--479},
  year={2024},
  organization={PMLR}
}
% Author, F.: Article title. Journal \textbf{2}(5), 99-110 (2016)

% F., Editor, S. (eds.) CONFERENCE 2016, LNCS, vol. 9999, pp. 1-13.


% on Proceedings, pp. 1-2. Publisher, Location (2010)

% LNCS Homepage, \url{http://www.springer.com/lncs}, last accessed 2023/10/25
% \end{thebibliography}
\end{document}